\PassOptionsToPackage{table}{xcolor}
\documentclass[lettersize,journal]{IEEEtran}
\usepackage{amsmath,amsfonts}
\usepackage{algorithmic}
\usepackage{algorithm}
\usepackage{array}
\usepackage[caption=false,font=normalsize,labelfont=sf,textfont=sf]{subfig}
\usepackage{textcomp}
\usepackage{stfloats}
\usepackage{url}
\usepackage{graphicx}
\usepackage{cite}
\usepackage{changes}
\usepackage{makecell}
\usepackage{graphics}   
\usepackage{epsfig}     
\usepackage{mathptmx}   
\usepackage{times}      
\usepackage{amssymb}    
\usepackage{booktabs}   
\usepackage{multirow}   
\usepackage{footnote}
\usepackage{color}
\usepackage{threeparttable}

\begin{document}

\title{
    \LARGE \bf
    E-S2Feat:Semantic-Guided Spiking Local Feature Detection and Description for Event Cameras
}

\author{
    Yang Yi, Juntao Hua, Jinpu Zhang, Liangwei Fan$^{*}$, Hui Shen$^{**}$ and Dewen Hu
    \thanks{
        Y. Yi, J. Hua, J. Zhang, L. Fan, H. Shen and D. Hu are with the College of Intelligence Science and Technology, National University of Defense Technology, China.
    }
    \thanks{
        ** Principal corresponding author : H. Shen(shenhui@nudt.edu.cn)
    }
    \thanks{
        * Corresponding author : Liangwei Fan(fanliangwei@nudt.edu.cn)
    }
    \thanks{
        This work has partially been funded by Research on Spatial Perception and Computing Technologies for Dynamic Complex Scenarios Based on Heterogeneous Brain-inspired Intelligence (U25B2069), Autonomous Learning Theory and Applications of Intelligent Unmanned Systems (T2521006) and Multi-modal Object Tracking in Complex Dynamic Scenarios (62506377).
    }
}

\maketitle
\thispagestyle{empty}
\pagestyle{empty}

\begin{abstract}
Benefiting from high temporal resolution and dynamic range, event-based local feature methods have attracted increasing attention. However, event sparsity, noise, and limited texture still hinder robust local feature learning. Deploying such methods on resource-constrained platforms such as unmanned aerial vehicles also requires balancing accuracy and energy efficiency.
To address these challenges, this paper proposes \textbf{E-S2Feat}, a spiking neural network framework for event-based local feature detection and description.
The framework jointly optimizes local feature learning from the perspectives of feature representation and selection.
First, a module-specific spiking activation mechanism preserves fine-grained structural cues and discriminative information under low-bit, energy-efficient inference, thereby improving overall representation fidelity.
Furthermore, a semantic-guided feature modulation mechanism leverages semantic priors to refine keypoint response distributions and enhance local descriptor discriminability, thereby guiding the model to extract local features with greater geometric stability and stronger discriminative capability.
Experiments on the ECD and EDS datasets show that the proposed method significantly outperforms baseline methods such as SuperEvent in pose estimation accuracy. It also achieves accuracy comparable to its artificial neural network counterpart while delivering an approximately 4.8-fold improvement in theoretical computational energy efficiency.
Visual-inertial odometry experiments on the TUM-VIE dataset further verify the effectiveness and practical application potential of the proposed method in complete SLAM systems.
\end{abstract}


\section{Introduction}
Local feature detection and description are fundamental tasks in computer vision. They provide essential support for downstream applications such as image matching~\cite{balntas2017hpatches, olson2002maximum}, 3D reconstruction~\cite{furukawa2009accurate, schonberger2018semantic}, and SLAM~\cite{qin2018vins, 11455506}. Local feature methods based on RGB images~\cite{lowe2004distinctive, detone2018superpoint, lindenberger2023lightglue, yi2026gessmulticueguidedlocal} have achieved strong performance in conventional scenarios. However, frame-based cameras are vulnerable to motion blur and exposure failure under high-speed motion or extreme lighting conditions. These effects often degrade feature quality and reduce the robustness of visual localization. In contrast, event cameras~\cite{lichtsteiner2008128} offer high temporal resolution, high dynamic range, and low power consumption~\cite{gallego2022event}. They can provide reliable visual information in scenes with rapid motion and challenging illumination. Event cameras have therefore shown strong potential in applications such as autonomous driving and visual SLAM~\cite{maqueda2018event, vidal2018ultimate}. Nevertheless, event streams are inherently sparse, noisy, and often lack texture, which can degrade the stability and repeatability of extracted features and thus compromise the performance of downstream visual localization.

In practical visual localization systems deployed on resource-constrained platforms such as unmanned aerial vehicles, the front-end feature extractor must not only ensure high matching accuracy but also meet real-time and energy-efficiency requirements, which motivates the exploration of more energy-efficient neural network computation.
In recent years, spiking neural networks (SNNs) have attracted increasing attention. SNNs encode information through spike timing and are generally considered more biologically plausible and energy-efficient than conventional artificial neural networks (ANNs). They have been applied to several event-based vision tasks, including classification~\cite{yao2021temporal}, object detection~\cite{luo2024integer}, and optical flow estimation~\cite{wan2022learning}. These studies demonstrate the potential of SNNs for low-power perception and event representation learning. However, existing learning-based methods for event feature detection and description still rely mainly on conventional ANNs architectures. 
Representative methods such as SuperEvent~\cite{burkhardt2025superevent} have demonstrated the effectiveness of architectures such as MaxViT~\cite{tu2022MaxViT} for event-based keypoint detection and description. These methods also achieve strong performance in pose estimation. However, they still rely on continuous-valued activations and conventional deep neural network computation.
In comparison, the spiking activation and event-driven computation of SNNs are more compatible with the asynchronous nature of event data. SNNs also support low-bit representations and energy-efficient inference~\cite{roy2019towards}. They therefore offer a promising way to improve the efficiency of event-based visual front-end while preserving feature discriminability.

\begin{figure}[t]
    \centering
    \includegraphics[width=0.48\textwidth]{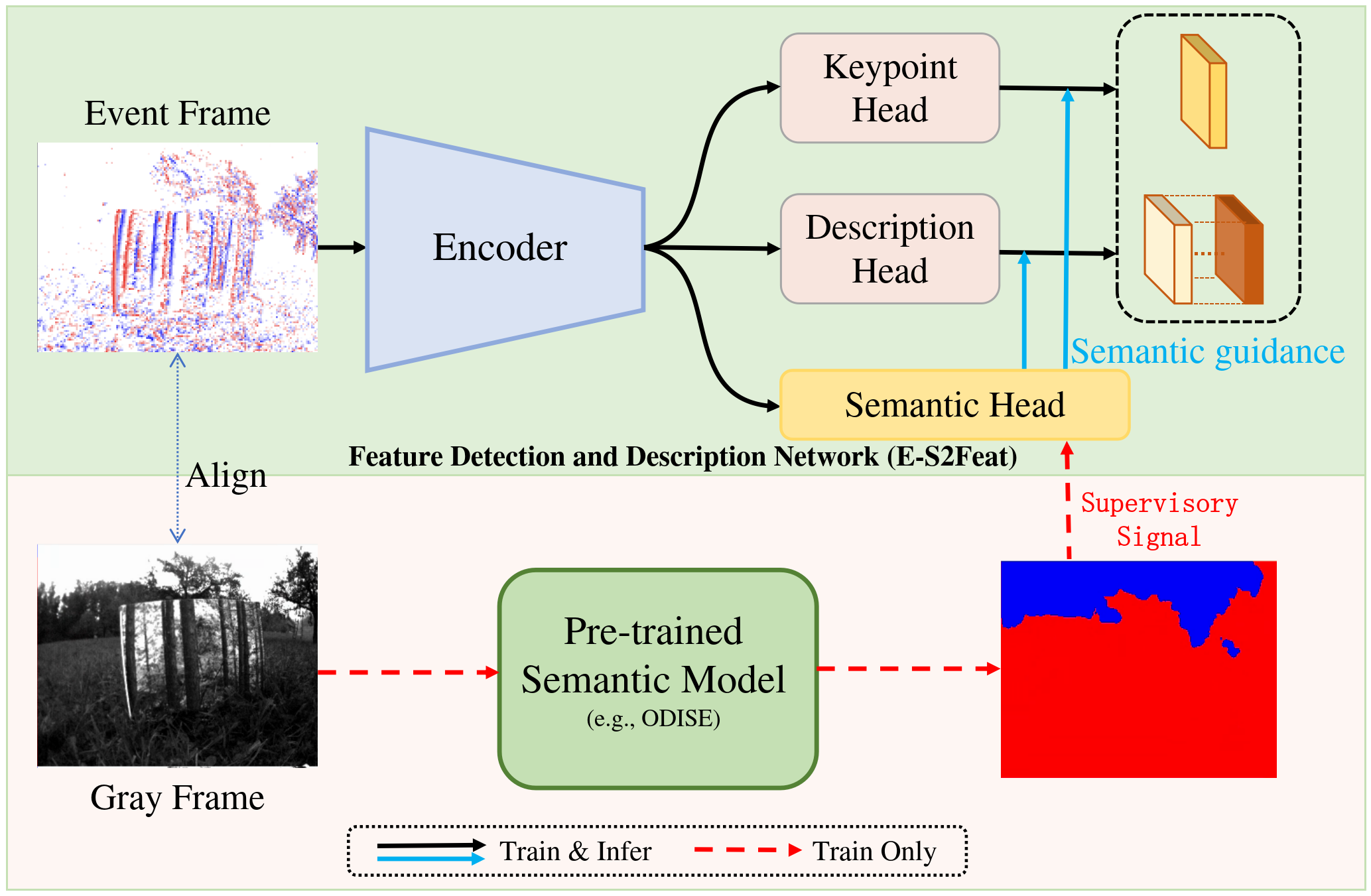}
    \caption{Overview of the proposed framework. The network is built upon spiking neural networks and incorporates a module-specific spiking activation mechanism, achieving low-bit, energy-efficient inference while preserving feature representation fidelity. Event frames are encoded by a shared spiking backbone and routed to the keypoint, descriptor, and semantic branches. The semantic branch jointly modulates keypoint responses and descriptor features to extract geometrically stable and discriminative local features. It is supervised by a pre-trained semantic model during training, which is unnecessary at inference time.}
    \label{fig:overview}
\end{figure}

We propose \textbf{E-S2Feat}, the first SNN-based framework for the joint detection and description of local features from event cameras. To improve the representational fidelity of local features, we design module-specific spiking strategies tailored to individual nonlinear modules. These strategies enable low-bit, energy-efficient inference while preserving the fine-grained structural details and discriminative cues essential for keypoint localization and descriptor learning. However, sufficiently faithful feature representation does not necessarily imply that all extracted features can provide stable geometric constraints. Regions such as the sky lack stable and repeatable local structures, where event noise can readily produce spurious patterns and induce unstable feature responses, thereby compromising the reliability of feature matching and pose estimation~\cite{gallego2022event}. To address this issue, we further introduce a semantically guided feature modulation mechanism. To the best of our knowledge, this is the first work to incorporate semantic priors into the joint detection and description of local features from event cameras. The proposed mechanism introduces complementary scene priors at the feature selection level. It exploits semantic information to identify regions that lack stable geometric structures, such as the sky, suppresses unreliable feature responses induced by event noise in these regions, and enhances local descriptor features in geometrically reliable regions. Specifically, we attach a lightweight semantic segmentation head to the shared backbone to predict semantic masks for geometrically unreliable regions. The predicted masks are then used to guide the detection and description branches through score modulation and feature fusion, respectively. Unlike conventional denoising methods that rely on spatio-temporal neighborhood correlations or event statistics~\cite{khodamoradi,guo2022low}, the proposed mechanism leverages high-level semantic information to optimize the distribution of keypoint responses and enhance the discriminative capability of local descriptors. In this way, it guides the model toward extracting local features that are more geometrically stable and discriminative.

Consequently, the module-specific spiking strategies and the semantically guided feature modulation mechanism jointly optimize event-based local feature learning at the feature representation and feature selection levels, respectively, together forming a unified framework for event-based local feature detection and description. The overall architecture of E-S2Feat is illustrated in Fig.~\ref{fig:overview}. In summary, the main contributions of this work are as follows:

\begin{enumerate}[]
\item We propose E-S2Feat, the first SNN-based framework for joint local feature detection and description using event cameras. It employs module-specific spiking strategies for different nonlinear components, enabling efficient low-bit inference while largely preserving feature representation capacity.
\item We introduce a semantically guided feature modulation mechanism into event-based local feature detection and description, guiding the model to extract local features that are more geometrically stable and discriminative.
\item Experiments on the ECD, EDS, and TUM-VIE datasets demonstrate that E-S2Feat achieves significant accuracy improvements in pose estimation and Visual-Inertial Odometry (VIO) tasks.
\end{enumerate}

\section{Related Works}
\subsection{Event-Based Local Feature Extraction}
Early research on local feature extraction for event cameras predominantly relied on handcrafted methods, including corner detection based on speed-invariant time surfaces~\cite{manderscheid2019speed}, as well as feature tracking and geometric estimation techniques specifically designed for event streams~\cite{alzugaray2018asynchronous,vidal2018ultimate,ikura2024rate}. These approaches are generally characterized by strong interpretability and low computational cost. However, their performance often depends heavily on manually engineered event representations and carefully tuned parameters, which limits their robustness and generalization in the presence of substantial noise, rapid motion, and complex scene variations~\cite{gallego2022event}.

In recent years, learning-based approaches to event-based local feature extraction have attracted growing attention. EventPoint~\cite{huang2023eventpoint} and SD2Event~\cite{10655335} jointly learn event-based keypoint detection and description in a self-supervised manner, while LLAK~\cite{chiberre2022long} focuses on improving the detection and tracking stability of long-lived keypoints in event streams. More recently, SuperEvent~\cite{burkhardt2025superevent} employs cross-modal supervision to learn robust event-based keypoints and descriptors and incorporates them into a keypoint-based SLAM framework, highlighting the potential of learned event local features for visual localization and pose estimation.

To the best of our knowledge, no prior study has explored SNNs for the joint detection and description of event-based local features. This research gap may be attributed to a fundamental challenge: local feature detection and description require the preservation of fine-grained spatial structures and highly discriminative local appearance cues~\cite{wang2023attention}, whereas the discrete and capacity-limited spike representations of conventional SNNs can introduce a severe information bottleneck. As a result, it remains difficult to simultaneously achieve the precise spatial localization required for keypoint detection and the rich representational capacity needed for robust feature description.

\subsection{Spiking Neural Networks for Vision}
SNNs encode information through the temporal dynamics of discrete spikes. Compared with conventional ANNs, they offer greater biological plausibility and potentially higher energy efficiency, and are widely regarded as a core computational paradigm for neuromorphic computing~\cite{tavanaei2019deep}. The advent of surrogate gradient methods~\cite{neftci2019surrogate} has enabled the end-to-end training of SNNs via backpropagation, substantially accelerating the development of deep SNNs. Nevertheless, conventional spiking neurons typically communicate through binary spikes~\cite{hodgkin1952quantitative,shaban2021adaptive,zheng2024temporal}, which restricts the amount of information that can be conveyed within a single time step. To compensate for this limited information capacity, high-performance SNNs often rely on multi-timestep rate coding, in which spike responses are accumulated over time to enrich feature representations. This strategy, however, incurs increasing inference latency and computational overhead as the number of simulation timesteps grows. To alleviate this limitation, Fan et al.~\cite{fan2025multisynaptic} introduced the Multi-Synaptic Firing (MSF) neuron, which employs multiple firing thresholds to generate multi-level discrete spike responses within a single time step. By increasing the information-carrying capacity of spiking neurons, MSF reduces their reliance on long temporal coding windows.

Despite this progress, multi-level spiking mechanisms such as MSF are primarily tailored to Rectified Linear Unit (ReLU) activations. Smooth nonlinearities widely used in modern visual backbones, including the Gaussian Error Linear Unit (GELU) and the Sigmoid Linear Unit (SiLU), still lack effective spiking counterparts capable of preserving their nonlinear response characteristics while maintaining computational efficiency. Moreover, existing SNNs research has largely focused on high-level visual tasks such as image classification~\cite{yao2021temporal} and object detection~\cite{luo2024integer}, whereas the potential of SNNs for event-based local feature detection and description remains largely unexplored. To address these limitations, we develop module-specific spiking strategies for heterogeneous nonlinear modules, assigning each module a spiking activation mechanism tailored to its functional characteristics. The proposed design aims to reduce the feature degradation caused by spike quantization while achieving a favorable balance between representation fidelity and computational efficiency.

\subsection{Semantics-Guided Feature Selection}
Existing methods commonly leverage semantic segmentation to exclude regions that are unlikely to provide reliable geometric constraints, such as dynamic objects and the sky, thereby reducing mismatches and invalid observations. For example, DynaSLAM~\cite{bescos2018dynaslam} and DS-SLAM~\cite{yu2018ds} exploit scene semantics to identify dynamic regions and improve localization robustness in complex environments. Beyond explicit feature filtering, semantic information has also been increasingly incorporated into local feature detection and description. SFD2~\cite{xue2023sfd2} introduces semantics-aware constraints to guide the detector toward stable regions, such as buildings and lane markings, while enhancing the discriminative capability of local descriptors. More recently, SAMFeat~\cite{wu2025segment} and GESS~\cite{yi2026gessmulticueguidedlocal} have further investigated the use of semantic priors in local feature extraction, description, and matching. Collectively, these studies demonstrate that semantic information can complement low-level textural and geometric cues with high-level scene priors, thereby facilitating the selection of stable features and the learning of more discriminative representations.

However, semantic-guided feature selection remains insufficiently explored for event-based local feature detection and description. Incorporating semantic priors can suppress unreliable feature responses and guide the model to focus on scene regions with stable geometric structures. To this end, this paper integrates a semantic branch into an end-to-end SNNs framework. Built upon the module-specific spiking activation mechanism that improves the representation fidelity of local features, the proposed semantic branch leverages the generated semantic masks to refine keypoint response distributions and enhance the discriminative capability of local descriptors, thereby enabling the extraction of local features with stronger geometric stability and discriminative power.

\section{METHODOLOGY}
E-S2Feat is an end-to-end SNN-based framework for local feature detection and description with event cameras. The framework jointly optimizes event-based local feature learning at both the feature representation and feature selection levels. At the representation level, the module-specific spiking strategy assigns appropriate spiking activation functions to different network modules according to their functional roles and feature distributions, thereby balancing feature representation capability and computational efficiency. At the selection level, semantic-guided feature modulation exploits scene-level semantic priors to constrain keypoint detection and descriptor generation, improving the geometric reliability and matching stability of local features. The following sections describe the overall architecture, module-specific spiking strategy, semantic-guided feature modulation mechanism, and training strategy.

\subsection{Overall System Architecture}
\begin{figure}[t]
    \centering
    \includegraphics[width=0.49\textwidth]{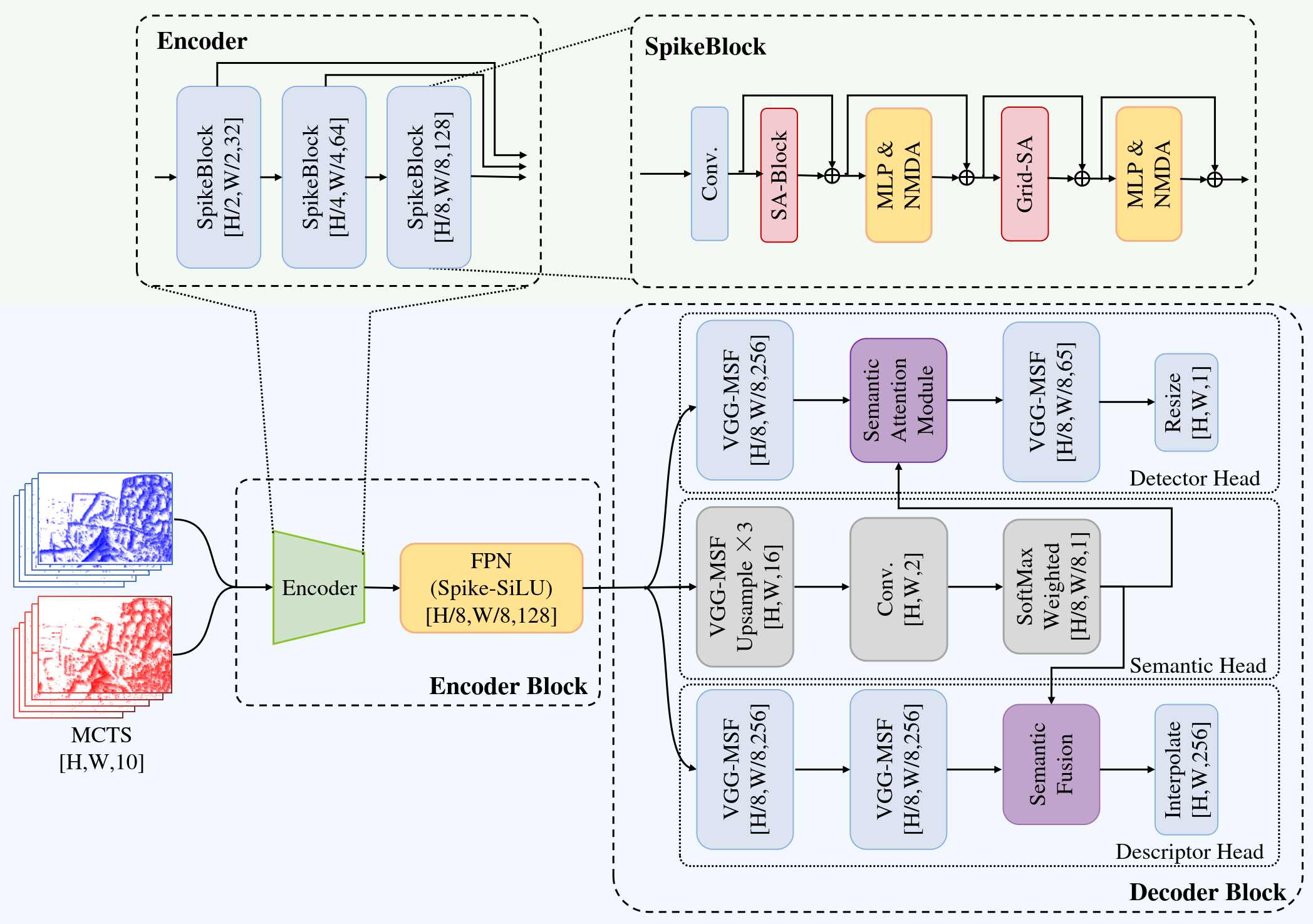}
    \caption{Overall architecture of E-S2Feat. The spiking MaxViT and FPN extract multi-scale features, while three task branches generate keypoint heatmaps, local descriptors, and a geometric reliability mask. The mask guides feature detection and description through score modulation and feature fusion.}
    \label{fig:network}
\end{figure}

The overall architecture of the proposed system is illustrated in Fig.~\ref{fig:network}. The input event stream is first converted into a Multi-Channel Time Surfaces (MCTS) representation~\cite{burkhardt2025superevent}. Specifically, five approximately logarithmically spaced temporal windows $\Delta t \in \{1,3,10,30,100\}$ ms and two polarity channels are used to construct a 10-channel multi-scale event representation. This representation preserves both short-term event dynamics and structural information over longer temporal intervals. The MCTS input is then hierarchically encoded by a spiking MaxViT backbone~\cite{tu2022MaxViT}. A feature pyramid network (FPN) further fuses event features across different spatial scales to produce a shared multi-scale representation. The fused features are fed into the detection, description, and semantic prediction heads, which generate a keypoint response heatmap, dense 256-dimensional local descriptors, and a geometric reliability mask, respectively. The semantic predictions guide keypoint detection and descriptor generation through score modulation and feature fusion. Finally, the system outputs semantically constrained keypoint locations and their corresponding local descriptors.

\subsection{Module-Specific Spiking Strategy}
A classical approach for converting an artificial neural network into a spiking neural network is to preserve the original topology and weights while replacing continuous activations, such as ReLU, with integrate-and-fire neurons. The corresponding analog activation values are then approximated by spike firing rates~\cite{rueckauer2017conversion}. However, within a limited number of time steps, discrete spike trains cannot precisely represent continuous activations. This mismatch introduces truncation and quantization errors, which may accumulate layer by layer during network propagation~\cite{yao2025towards}. In local feature extraction networks, different modules serve different purposes. Attention modeling, multi-scale feature fusion, and dense semantic prediction have distinct activation distributions, dynamic ranges, and information representation requirements. Previous studies have shown that spike quantization errors are closely related to membrane potential distributions. Applying a uniform spiking activation across different modules may therefore exacerbate activation mismatch and information loss~\cite{guo2023rmp}.

To this end, we employ module-specific spiking designs for different network components. Specifically, spiking GELU and spiking SiLU are developed to replace the original GELU and SiLU activations, respectively, while ReLU is replaced with the MSF multi-level spiking activation proposed in our previous work~\cite{fan2025multisynaptic}. By assigning each module a spiking activation tailored to its functional role and feature distribution, the proposed design preserves the sparsity of spike-based computation while reducing the adverse effects of discrete encoding on local feature detection and description performance.

\subsubsection{Assignment of Spiking Activations}
\paragraph{\textbf{MSF Multi-Level Spiking Activation}}
The detection, description, and semantic heads all require preserving the hierarchy of feature magnitudes within a single time step. However, conventional binary spiking neurons can only represent two states, i.e., spike or no spike, which inevitably leads to excessive compression of feature information. To address this limitation, we replace the ReLU activations in the network with MSF neurons and specifically employ them in the detection, description, and semantic heads to enhance each task branch's capacity to represent fine-grained feature magnitudes. An MSF neuron consists of $D$ synaptic branches with different firing thresholds. For the $d$-th branch, a binary spike is generated as $s^d=\Theta\left(u_i-V_{\mathrm{th}}^d\right)$ when the membrane potential $u_i$ reaches the corresponding threshold $V_{\mathrm{th}}^d$, where $\Theta(\cdot)$ denotes the Heaviside step function. By aggregating the output spikes from all branches, the MSF neuron can generate an integer spike count ranging from $0$ to $D$ within a single time step, thereby providing $D+1$ discrete magnitude levels. The corresponding neuronal dynamics follow those described in~\cite{fan2025multisynaptic}.

For the $l$-th ReLU activation, let $x_{\max}^{(l)}$ denote the calibrated upper bound of the corresponding activation values. Given an MSF neuron with $D$ synaptic branches, the quantization step size at this activation is defined as $\Delta_l=x_{\max}^{(l)}/D$. To ensure that the spike count of the MSF neuron corresponds to a round-to-nearest quantization, the firing threshold of the $d$-th synaptic branch is set to $V_{\mathrm{th}}^{(l,d)}=\left(d-1/2\right)\Delta_l$. Consequently, the integer spike count generated by the MSF neuron can be expressed as:
\begin{equation}
    n_D(x)=\operatorname{clip}\!\left(
    \left\lfloor\frac{x}{\Delta_l}+\frac{1}{2}\right\rfloor, 0,D\right).
\end{equation}
The corresponding reconstructed output is given by:
\begin{equation}
    \widehat{x}_D=\Delta_l \cdot n_D(x)
\end{equation}
where $\operatorname{clip}(\cdot,0,D)$ constrains the integer spike count to the range $[0,D]$. For inputs within the calibrated range without upper-bound saturation, i.e., $x\in[0,x_{\max}^{(l)}]$, the reconstruction error of the above round-to-nearest quantization satisfies:
\begin{equation}
    \left|\widehat{x}_D-x\right|
    \leq\frac{\Delta_l}{2}
    =\frac{x_{\max}^{(l)}}{2D}.
\end{equation}
Therefore, with a fixed $x_{\max}^{(l)}$, the upper bound of the quantization error decreases at a rate of $1/D$ as the number of synaptic branches $D$ increases, indicating that a larger number of branches enables a finer approximation of feature magnitudes.

Fan et al.~\cite{fan2025multisynaptic} showed that setting $D=4$ achieves a favorable trade-off between accuracy and energy efficiency for classification and related tasks. Under this configuration, both the integer spike-count range and the representation overhead remain relatively low, making this configuration suitable for tasks with modest requirements on feature magnitude resolution. In contrast, the semantic branch in our framework is required not only to predict regional semantic categories but also to generate pixel-level semantic responses for subsequent keypoint score modulation and descriptor feature fusion. Motivated by these requirements, we uniformly set the number of branches of all MSF neurons to $D=32$ in our experiments. This configuration substantially reduces the quantization step size, thereby preserving finer-grained spatial responses and feature magnitude hierarchies under single-time-step inference.

\paragraph{\textbf{Spiking GELU Activation}} 
The attention MLP in MaxViT~\cite{tu2022MaxViT} relies on the GELU activation to perform nonlinear feature transformation along the channel dimension, making its output particularly sensitive to activation approximation errors. To reduce the activation mismatch introduced by spike-based conversion, we propose Spiking GELU, which approximates the GELU response using a continuous basis expansion with the correct wide-range asymptotic behavior. For an input $x$, the continuous approximation is defined as
\begin{equation}
    Z_{\mathrm{wide}}(x)
    =w_0x+c\sqrt{x^2+\delta}+b_0
    +\sum_{j=1}^{B}\kappa_jH\!\left(\frac{x}{\sigma_j}\right)
\end{equation}
where $H(u)=(1-u^2)\cdot\exp(-u^2/2)$ denotes the Mexican-hat wavelet basis~\cite{ricker1951form}, $\kappa_j$ and $\sigma_j$ represent the amplitude coefficient and scale parameter of the $j$-th wavelet basis, respectively, $B$ is the number of wavelet bases, and $\delta$ is introduced to smooth the square-root basis function. Accordingly, the asymptotic slopes of the continuous approximation at positive and negative infinity are $w_0+c$ and $w_0-c$, respectively. To match the asymptotic behavior of GELU, which approaches the linear function $x$ in the positive region and zero in the negative region, we set $w_0=c=0.5$ and constrain $b_0$ to be close to zero during parameter fitting. Consequently, the linear term together with the even-symmetric square-root term captures the wide-range asymptotic structure of GELU, while the wavelet bases with localized support are employed to fit the residual decay in the intermediate activation range, thereby reducing the tail-slope mismatch caused by finite local basis expansions.

The continuous approximation is subsequently discretized using signed multisynaptic spiking neuron. To this end, we define the symmetric round-to-nearest integer operator as:
\begin{equation}
    \mathcal{R}(z)=
    \operatorname{sgn}(z)
    \left\lfloor |z|+\frac{1}{2}\right\rfloor.
\end{equation}
The integer spike output and the corresponding reconstructed output of Spiking GELU are respectively given by:
\begin{equation}
    \left\{
    \begin{aligned}
    s_{\mathrm{GELU}}(x)&=\operatorname{clip}\!\left(\mathcal{R}\!\left(\frac{Z_{\mathrm{wide}}(x)}{\gamma}\right),-m_G,M_G\right),\\
    \widehat{y}_{\mathrm{GELU}}(x)&=\gamma \cdot s_{\mathrm{GELU}}(x)
    \end{aligned}
    \right.
\end{equation}
where $\gamma$ represents the output quantization step size, while $M_G$ and $m_G$ specify the maximum positive and negative integer spike counts of Spiking GELU, respectively. Based on the negative dynamic range of GELU and the calibrated positive output bound $y_{\max}$, they are determined as:
\begin{equation}
    \left\{
    \begin{aligned}
    m_G&=\max\!\left(1,\left\lceil\frac{\varepsilon_G}{\gamma}\right\rceil\right), \\
    M_G&=\left\lceil\frac{y_{\max}}{\gamma}\right\rceil
    \end{aligned}
    \right.
\end{equation}
where $\varepsilon_G\triangleq-\min_x\operatorname{GELU}(x)\approx 0.17$ denotes the magnitude of the minimum value of the ideal GELU function in the negative region. Since the negative responses of GELU have relatively small magnitudes, a limited negative spike range is sufficient to cover its principal negative responses, whereas the positive spike range is determined by the actual activation distribution. In this work, we set the quantization step size to $\gamma=0.1$, yielding $m_G=2$ and a negative representation lower bound of $-0.2$, which fully covers the negative responses of GELU.

\paragraph{\textbf{Spiking SiLU Activation}}
The SiLU activation in the FPN convolutional modules is approximated by the proposed Tail-Corrected Wavelet (TCW) formulation for spiking conversion. For an approximation function composed of a linear term, a constant term, and a finite number of wavelet bases, the wavelet components decay to zero as $|x|\rightarrow\infty$, causing the tail behavior to be ultimately dominated by the linear term. Therefore, simply increasing the number of wavelet bases, reducing the quantization step size, or expanding the clipping range cannot eliminate the tail errors caused by mismatched asymptotic structures. To preserve the bilateral asymptotic properties of the SiLU function, we adopt the following decomposition:
\begin{equation}   
    \left\{
    \begin{aligned}
    &\operatorname{SiLU}(x) = \operatorname{ReLU}(x)+r(|x|), \\
    &r(u)=-u\cdot\operatorname{sigmoid}(-u), \quad u\geq 0
    \end{aligned}
    \right.
\end{equation}
where the residual function $r(u)\in[-\varepsilon_S,0]$ approaches zero as $u\rightarrow\infty$, and $\varepsilon_S\triangleq-\min_u r(u)\approx 0.278465$ denotes the magnitude of the minimum negative value of SiLU. In this work, only this bounded decaying residual is approximated using wavelet bases:
\begin{equation} 
    \widehat{r}(u) = \sum_{j=1}^{B} \kappa_j H\!\left( \frac{u-\mu_j}{\sigma_j} \right)
\end{equation}
where $\mu_j$ denotes the center of the $j$-th wavelet basis. The corresponding tail-corrected continuous output is defined as:
\begin{equation} 
    Z_{\mathrm{TCW}}(x) = \operatorname{ReLU}(x) + \widehat{r}(|x|). 
\end{equation} 
No constant bias term is introduced in the residual approximation to ensure $\widehat{r}(u)\rightarrow 0$. Meanwhile, the constraint $\widehat{r}(0)=0$ is imposed during parameter optimization to preserve the value of SiLU at the origin. Accordingly, the wavelet parameters are obtained through the following constrained offline fitting process:
\begin{equation} 
    \begin{aligned} 
        \min_{\{\kappa_j,\mu_j,\sigma_j\}_{j=1}^{B}} \quad & \int_{0}^{U} \left[ \widehat{r}(u) + u\,\operatorname{sigmoid}(-u) \right]^2 \,\mathrm{d}u, \\ \text{s.t.} \quad & \widehat{r}(0)=0, \qquad \sigma_j>0. 
    \end{aligned} 
\end{equation}
In this work, a fixed set of $B=3$ wavelet parameters is fitted offline within the interval $[0,16]$ and shared across all SiLU activation sites. Therefore, the proposed method does not require storing additional independent fitting parameters for different activation locations. During the discrete spiking stage, the symmetric round-to-nearest integer operator $\mathcal R(\cdot)$ defined previously is reused to quantize the continuous output as
\begin{equation}
    \left\{
    \begin{aligned}
        s_{\mathrm{SiLU}}(x) &= \operatorname{clip}\!\left( \left\lfloor \frac{Z_{\mathrm{TCW}}(x)}{\gamma} +\frac{1}{2} \right\rfloor, -m_S, M_S \right), \\
        \widehat{y}_{\mathrm{SiLU}}(x) &= \gamma\ \cdot s_{\mathrm{SiLU}}(x)
    \end{aligned}
    \right.
\end{equation}
where $\gamma$ denotes the output quantization step size, and $M_S$ and $m_S$ define the positive and negative integer spike ranges of Spiking SiLU, respectively:
\begin{equation}
    \left\{
    \begin{aligned}
        m_S &= \max\!\left( 1, \left\lceil \frac{\varepsilon_S}{\gamma} \right\rceil \right), \\
        M_S &= \left\lceil \frac{x_{\max}}{\gamma} \right\rceil
    \end{aligned}
    \right.
\end{equation}
where $x_{\max}$ represents the calibrated upper bound of the positive output. The negative spike range $m_S$ is determined according to the theoretically defined minimum magnitude of SiLU, $\varepsilon_S$, rather than adopting the negative range used for GELU. This design covers the negative dynamic range of SiLU and avoids additional saturation errors caused by negative clipping.

\subsubsection{Single-Step Spike Encoding and Inference}
In this work, the network is configured with a single time step. Unlike conventional spike encoding schemes that accumulate information through firing rates over multiple time steps, the input MCTS representation~\cite{burkhardt2025superevent} has already explicitly encoded the multi-scale temporal information of the event stream using multiple temporal windows. Therefore, it is unnecessary to further unfold the temporal dimension within the network. Single-step inference avoids the latency and computational overhead associated with multi-step forward propagation, making it better suited to the real-time requirements of local feature extraction from event data. Accordingly, each spiking activation maps continuous features into finite-level discrete responses. After calibration and quantization, the activations are represented as low-bit spike counts or integer states, enabling subsequent linear operations to be implemented as integer- or spike-based accumulations, thereby reducing storage overhead and theoretical computational energy consumption.

In summary, the proposed module-specific spiking activations together with the single-step discrete encoding improve the ability of different functional modules to approximate continuous nonlinearities while effectively controlling inference latency and computational cost. Nevertheless, the limited number of discrete response levels inevitably compresses feature magnitudes and fine-grained discriminative information, whereas increasing either the number of time steps or the response levels incurs additional inference latency and computational overhead. To address this limitation, rather than compensating for information loss solely by improving the precision of spike representations, we further introduce a semantic-guided feature modulation mechanism from the perspective of feature selection. This mechanism exploits semantic information to identify geometrically unreliable regions, such as the sky, and adaptively modulates keypoint responses while enhancing local descriptor features, thereby guiding the network to extract geometrically more stable and discriminative local features. Consequently, the proposed module-specific spiking activations and the semantic-guided mechanism collaboratively optimize event-based local feature learning from the complementary perspectives of feature representation and feature selection, respectively.

\subsection{Semantics-guided Feature Modulation}
Visual pose estimation is typically formulated under the assumption of a static and rigid scene, where the inter-frame motion of static feature points is induced primarily by camera motion~\cite{bescos2018dynaslam}. However, outdoor scenes often contain several types of geometrically unreliable regions. The sky and nearby distant regions lack stable structures, contain limited texture information, and exhibit high depth uncertainty, making them unsuitable for providing reliable geometric constraints. Dynamic vehicles and pedestrians undergo independent motion and therefore cannot be directly used for camera pose estimation~\cite{9812345}. Consequently, even when events in these regions are triggered by genuine brightness changes, they may still constitute structured interference from the perspective of pose estimation.

Motivated by these observations, we introduce semantic priors into the local feature learning process. For semantic category design, SFD2~\cite{xue2023sfd2} employs multi-class semantic priors to assign different weights to different regions. However, because the sparsity of event data makes fine-grained semantic categories difficult to predict reliably, we adopt a binary partition that distinguishes geometrically unreliable regions from reliable background regions and assigns them weights of 0 and 1, respectively. The sky can be identified relatively reliably using cues such as brightness and spatial location. In contrast, dynamic vehicles and pedestrians are more difficult to recognize because grayscale inputs lack color-discriminative cues and the available training data remain limited, leading to a relatively high false-negative rate. Future work will incorporate larger-scale annotations of dynamic scenes to improve dynamic-object recognition and explore finer-grained semantic modeling.

Based on this binary semantic partition, we generate semantic masks for geometrically unreliable regions and use them to guide the detection and descriptor branches through score modulation and feature fusion, respectively. The two mechanisms are described in the following subsections.

\subsubsection{Keypoint Response Modulation}
This part corresponds to the semantic attention module in Fig.~\ref{fig:network}. Specifically, we first apply spatial rearrangement-based downsampling to the semantic probability map so that its spatial resolution matches that of the detection features. The resulting representation is then processed by two mapping layers, each consisting of a convolution, batch normalization, and ReLU activation, followed by a Sigmoid function to generate a spatial attention map $W_s \in [0,1]^{H/8 \times W/8}$. The attention map modulates the detection features in a residual manner:
\begin{equation}
F_{\text{det}}' = F_{\text{det}} \odot \bigl(1 + \alpha \cdot W_s\bigr).
\end{equation}
where $F_{\text{det}}$ and $F_{\text{det}}'$ denote the detection features before and after modulation, respectively, and $\odot$ represents element-wise multiplication. The learnable scaling parameter $\alpha$, initialized to 0.5, adaptively controls the strength of semantic modulation.

Under this residual modulation scheme, attention weights in geometrically unreliable regions approach zero, leaving the corresponding feature responses largely unchanged. In contrast, geometrically reliable regions receive larger enhancement weights and therefore produce stronger keypoint responses. During the subsequent Top-K selection, candidate keypoints located in geometrically stable regions are preferentially retained. This encourages the detector to reduce its responses in unreliable regions and focus on scene structures that can provide stable geometric constraints.

\subsubsection{Descriptor-Semantic Feature Fusion}
This part corresponds to the semantic fusion module in Fig.~\ref{fig:network}. The module adopts a three-branch parallel architecture consisting of a semantic encoding branch, a channel attention branch, and a semantic guidance branch. The semantic encoding branch maps the input semantic map into a 48-dimensional semantic feature $F_{\mathrm{sem}}$ through two $3\times3$ convolutional layers. The channel attention branch employs the SE mechanism~\cite{hu2018senet}. It first aggregates the global statistics of the descriptor features through global average pooling and then generates the channel attention weights $W_{\mathrm{ch}}$ using a two-layer fully connected network with a reduction ratio of 8. The semantic guidance branch maps the semantic feature to a spatial attention weight $W_{\mathrm{sp}}$ through a $1\times1$ convolution, thereby encoding the geometric reliability of different spatial locations. The outputs of the three branches are integrated through joint channel-spatial modulation and residual fusion:
\begin{equation}
\begin{aligned}
    D_m &= D \odot W_{ch} \odot W_{sp}, \\
    D_f &= \operatorname{Conv}_{3\times3}(\operatorname{Concat}[D_m, F_{sem}]), \\
    D' &= D + \beta \cdot D_f.
\end{aligned}
\end{equation}
where $D$ and $D'$ denote the original descriptor feature and the semantically enhanced descriptor feature, respectively, and $D_m$ denotes the intermediate feature jointly modulated by the channel and spatial attention weights. The residual scaling coefficient $\beta$ is set to $0.5$ to control the contribution of semantic information during feature fusion. This residual fusion strategy supplements and enhances the original descriptor representation with scene-level semantic information, while preventing the semantic features from directly replacing the local texture representations learned by the backbone.

\subsection{Network Training}
\subsubsection{Supervision Signals}
The network is trained end to end under a multi-task supervision framework. The supervision signals are obtained from three sets of pseudo-labels. The pseudo-labels for keypoint detection and descriptor matching are jointly generated by pretrained SuperPoint~\cite{detone2018superpoint} and SuperGlue~\cite{sarlin2020superglue} using synchronized grayscale frames. Pixel-wise pseudo-labels for the semantic branch are generated by applying the open-vocabulary segmentation model ODISE~\cite{xu2023open} to the target datasets.

\subsubsection{Loss Functions}
During training, the detection and descriptor branches are supervised by the detection loss $L_{\text{det}}$ and descriptor loss $L_{\text{desc}}$, respectively, following the design of SuperEvent~\cite{burkhardt2025superevent}. Specifically, $L_{\mathrm{det}}$ is an $8\times8$ grid-based classification cross-entropy loss with an additional dustbin channel, whereas $L_{\mathrm{desc}}$ is formulated as a hinge loss. To provide explicit supervision for the newly introduced semantic branch, we further employ a pixel-wise semantic loss $L_{\mathrm{sem}}$. For each pixel $(i,j)$, let $p_{ij}$ denote the predicted probability that the pixel belongs to a geometrically unreliable region, $y_{ij}\in\{0,1\}$ denote the corresponding pseudo-label, and $|\Omega|$ denote the total number of pixels. The semantic loss is then defined as the pixel-wise binary cross-entropy:
\begin{equation}
L_{\text{sem}} = -\frac{1}{|\Omega|}\sum_{(i,j) \in \Omega} \bigl[ y_{ij} \log p_{ij} + (1-y_{ij}) \log(1-p_{ij}) \bigr].
\end{equation}
The total loss is a weighted sum of the three terms:
\begin{equation}
L_{\text{total}} = L_{\text{det}} + \lambda_{\text{desc}} L_{\text{desc}} + L_{\text{sem}}.
\end{equation}
where $\lambda_{\text{desc}}=10$ is the weight assigned to the descriptor loss.

\subsubsection{Data Augmentation}
During training, event-noise augmentation~\cite{gallego2022event} and homographic augmentation~\cite{detone2018superpoint} are applied jointly. Event-noise augmentation simulates common sensor effects, including contrast fluctuations, timestamp quantization errors, and defective pixels. Homographic augmentation applies random perspective transformations, scaling, and rotation to both the inputs and labels. This improves the robustness of the model to viewpoint changes.

\begin{table*}[t]
    \centering
    \footnotesize
    \caption{Pose Estimation Results on ECD\cite{mueggler2017event} and EDS~\cite{hidalgo2022event} datasets.}
    \setlength\tabcolsep{10pt}
    \renewcommand{\arraystretch}{1.4}
    \label{tab:sota}
    \begin{tabular}{c c c c c c c}
        \toprule
        \multirow{2}{*}[-3pt]{Methods} & \multicolumn{3}{c}{ECD: AUC [\%]} & \multicolumn{3}{c}{EDS: AUC [\%]} \\
        \cmidrule(lr){2-4} \cmidrule(lr){5-7}
        & @5$^\circ$ $\uparrow$& @10$^\circ$ $\uparrow$& @20$^\circ$ $\uparrow$& @5$^\circ$ $\uparrow$& @10$^\circ$ $\uparrow$& @20$^\circ$ $\uparrow$\\
        \midrule
        RATE~\cite{ikura2024rate} & 3.3 & 8.4 & 18.0 & 2.1 & 5.1 & 10.3 \\
        LLAK~\cite{chiberre2022long} & 0.7 & 1.4 & 2.1 & 0.5 & 0.7 & 10.3 \\
        SD2Event~\cite{gao2024sd2event} & 16.6 & 25.3 & 36.5 & 11.5 & 22.1 & 35.7 \\
        STPNet~\cite{zhu2025spatio} & 19.1 & 31.3 & 42.9 & 13.9 & 24.9 & 38.2 \\
        SuperEvent~\cite{burkhardt2025superevent} & \underline{22.8} & 35.8 & 46.7 & 25.4 & 37.5 & 49.0 \\
        \midrule
        \rowcolor{gray!20}
        \textbf{Ours (ANNs)} & \textbf{23.0} & \textbf{42.4} & \textbf{59.6} & \textbf{34.8} & \textbf{50.9} & \textbf{64.6} \\
        \rowcolor{gray!20}
        \textbf{Ours (SNNs)} & \underline{22.8} & \underline{42.1} & \underline{59.4} & \underline{34.1} & \underline{50.2} & \underline{64.0} \\
        \bottomrule
    \end{tabular}
\end{table*}

\section{EXPERIMENTS}
To validate the effectiveness of the proposed method, we conduct a comprehensive evaluation from five perspectives: pose estimation, ablation studies, qualitative visualization, VIO Trajectory Evaluation, and inference efficiency.

\subsection{Implementation}
The proposed method is implemented in PyTorch. Model training is performed on a single NVIDIA H100 GPU. All evaluation experiments are conducted on a separate workstation equipped with an NVIDIA RTX 4090 GPU. The training data configuration follows SuperEvent~\cite{burkhardt2025superevent}. We retain FPV~\cite{delmerico2019ready}, Griffin~\cite{rodriguez2021griffin}, and MVSEC~\cite{zhu2018multivehicle}. DDD20~\cite{hu2020ddd20} and ViViD++~\cite{lee2022vivid} in the original configuration are replaced with DSEC~\cite{gehrig2021dsec}. The semantic annotations in DDD20 and ViViD++ are of limited quality and cannot provide reliable supervision for the semantic branch. In contrast, DSEC provides higher-quality semantic annotations. For DSEC, the input is downsampled by a factor of 0.5, and the bottom 10 rows of pixels are masked out to remove sensor-edge artifacts. All inputs are cropped to resolutions divisible by the $8\times8$ grid size. The network is trained using the Adam optimizer with an initial learning rate of $10^{-3}$ and a batch size of 12. Training lasts for 10 epochs.

\subsection{Pose Estimation}
\subsubsection{Datasets}
The Event Camera Dataset~\cite{mueggler2017event} (ECD) is a standard benchmark for event-camera pose estimation. It has a resolution of $180\times240$ and contains multiple indoor and outdoor sequences with ground-truth 6-DoF camera poses. Event-aided Direct Sparse Odometry~\cite{hidalgo2022event} (EDS) is a higher-resolution event-camera pose estimation dataset with a resolution of $480\times640$. It was captured using a DAVIS sensor and contains more richly textured scenes, making pose estimation more challenging.

\subsubsection{Metrics}
Following the evaluation protocol in~\cite{burkhardt2025superevent}, we use the area under the curve (AUC) of the rotation error at thresholds of 5$^\circ$, 10$^\circ$, and 20$^\circ$ as the evaluation metrics. During evaluation, candidate keypoints are selected using a detection threshold of $\tau=0.01$. Fast Non-Maximum Suppression with a kernel size of $5\times5$ is then applied to remove duplicate responses. Pose estimation is performed using RANSAC thresholds of $t=1.0$ for ECD and $t=3.0$ for EDS.

\subsubsection{Results}
To ensure reliable evaluation, we corrected two potential issues identified in the original evaluation code. We then re-evaluated SuperEvent~\cite{burkhardt2025superevent} under the same experimental settings. Tab.~\ref{tab:sota} compares the pose estimation accuracy of different methods on the ECD and EDS datasets. Ours (ANNs) denotes the ANNs model equipped with semantic modulation, while Ours (SNNs) denotes its spiking counterpart. On ECD, Ours (ANNs) achieves an AUC@20$^\circ$ of 59.6\%, outperforming the re-evaluated baseline by 12.9 percentage points. Ours (SNNs) achieves 59.4\%, which is nearly identical to the ANNs version. On the higher-resolution EDS dataset, Ours (ANNs) and Ours (SNNs) achieve AUC@20$^\circ$ scores of 64.6\% and 64.0\%, respectively. Both models substantially outperform the other methods. Notably, the SNNs version remains comparable to the ANNs version across all metrics. This result shows that the proposed spiking design effectively preserves the model's ability to detect and describe event-based features.

\subsection{Ablation Studies}
To evaluate the contribution of each component, we conduct a progressive ablation study starting from the SuperEvent~\cite{burkhardt2025superevent} baseline. Components are added sequentially, and the results are reported in Tab.~\ref{tab:ablation}.
Starting from the original SuperEvent~\cite{burkhardt2025superevent} training configuration, we first evaluate the effect of introducing the DSEC dataset~\cite{gehrig2021dsec}. The AUC@20$^\circ$ increases from 46.7\% to 55.0\% on ECD and from 49.0\% to 60.8\% on EDS. These improvements indicate that training data quality has a substantial effect on event-based feature learning. We then introduce the semantic modulation module. The AUC@20$^\circ$ further increases to 59.6\% on ECD and 64.6\% on EDS. Finally, we construct the spiking version using the proposed module-specific spiking activations. Compared with the ANNs model, the performance decreases by only 0.2 percentage points on ECD, from 59.6\% to 59.4\%, and by 0.6 percentage points on EDS, from 64.6\% to 64.0\%. This minor accuracy loss confirms the effectiveness of the proposed module-specific spiking strategy.

\begin{table*}[t]
    \centering
    \footnotesize
    \caption{Ablation Experiment Results on ECD~\cite{mueggler2017event} and EDS~\cite{hidalgo2022event} datasets.}
    \setlength\tabcolsep{5pt}
    \renewcommand{\arraystretch}{1.4}
    \label{tab:ablation}
    \begin{tabular}{l c c c c c c}
        \toprule
        \multirow{2}{*}[-2pt]{Methods} & \multicolumn{3}{c}{ECD: AUC [\%]} & \multicolumn{3}{c}{EDS: AUC [\%]} \\
        \cmidrule(lr){2-4} \cmidrule(lr){5-7}
        & @5$^\circ$ $\uparrow$ & @10$^\circ$ $\uparrow$ & @20$^\circ$ $\uparrow$ & @5$^\circ$ $\uparrow$ & @10$^\circ$ $\uparrow$ & @20$^\circ$ $\uparrow$ \\
        \midrule
        SuperEvent~\cite{burkhardt2025superevent} & \underline{22.8} & 35.8 & 46.7 & 25.4 & 37.5 & 49.0 \\
        \rowcolor{gray!20}
        \quad + refined training data & 21.2 & 38.9 & 55.0 & 33.2 & 47.8 & 60.8 \\
        \rowcolor{gray!20}
        \quad + Semantic & \textbf{23.0} & \textbf{42.4} & \textbf{59.6} & \textbf{34.8} & \textbf{50.9} & \textbf{64.6} \\
        \rowcolor{gray!20}
        \quad + Spiking (\textbf{Ours SNNs}) & \underline{22.8} & \underline{42.1} & \underline{59.4} & \underline{34.1} & \underline{50.2} & \underline{64.0} \\
        \bottomrule
    \end{tabular}
\end{table*}

\subsection{Qualitative Visualization}
\begin{figure}[t]
    \centering
    \includegraphics[width=0.48\textwidth]{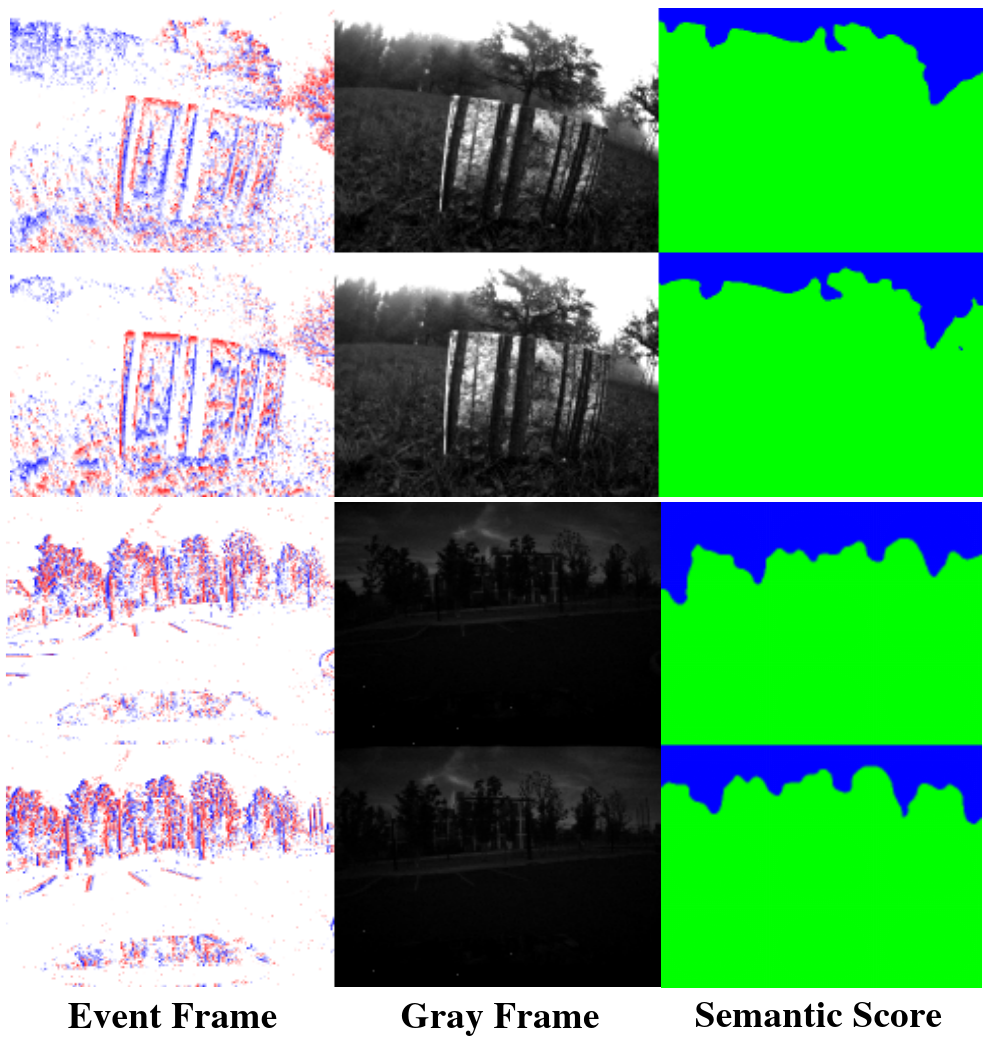}
    \caption{Visualization of the semantic segmentation output produced by Ours (SNNs). The three columns show the event frame, the grayscale frame, and the prediction from the SNNs semantic branch, respectively. Blue indicates geometrically unreliable regions, such as the sky, while green indicates valid background regions retained for local feature extraction.}
    \label{fig:semantic}
\end{figure}

Fig.~\ref{fig:semantic} presents representative outputs of the model’s semantic segmentation head, where blue denotes geometrically unreliable regions, such as the sky, and green indicates the retained valid background regions. Even under low-illumination and low-contrast conditions, the SNN-based semantic branch can clearly distinguish between the two types of regions and recover their structural boundaries, thereby providing effective semantic guidance for subsequent detection-score modulation and descriptor fusion. Although spike quantization may limit the representation of fine-grained boundary details, the semantic branch equipped with MSF-based multi-level spike activations still achieves prediction quality sufficient for semantic gating. Considering the trade-off among prediction difficulty, computational cost, and semantic guidance effectiveness, we adopt a binary segmentation scheme that distinguishes only between geometrically unreliable regions and valid background regions, rather than performing fine-grained multi-class segmentation. This binary prediction provides adequate geometric priors for feature modulation while maintaining a practical balance between model complexity and semantic guidance performance.

\begin{figure}[t]
    \centering
    \includegraphics[width=0.48\textwidth]{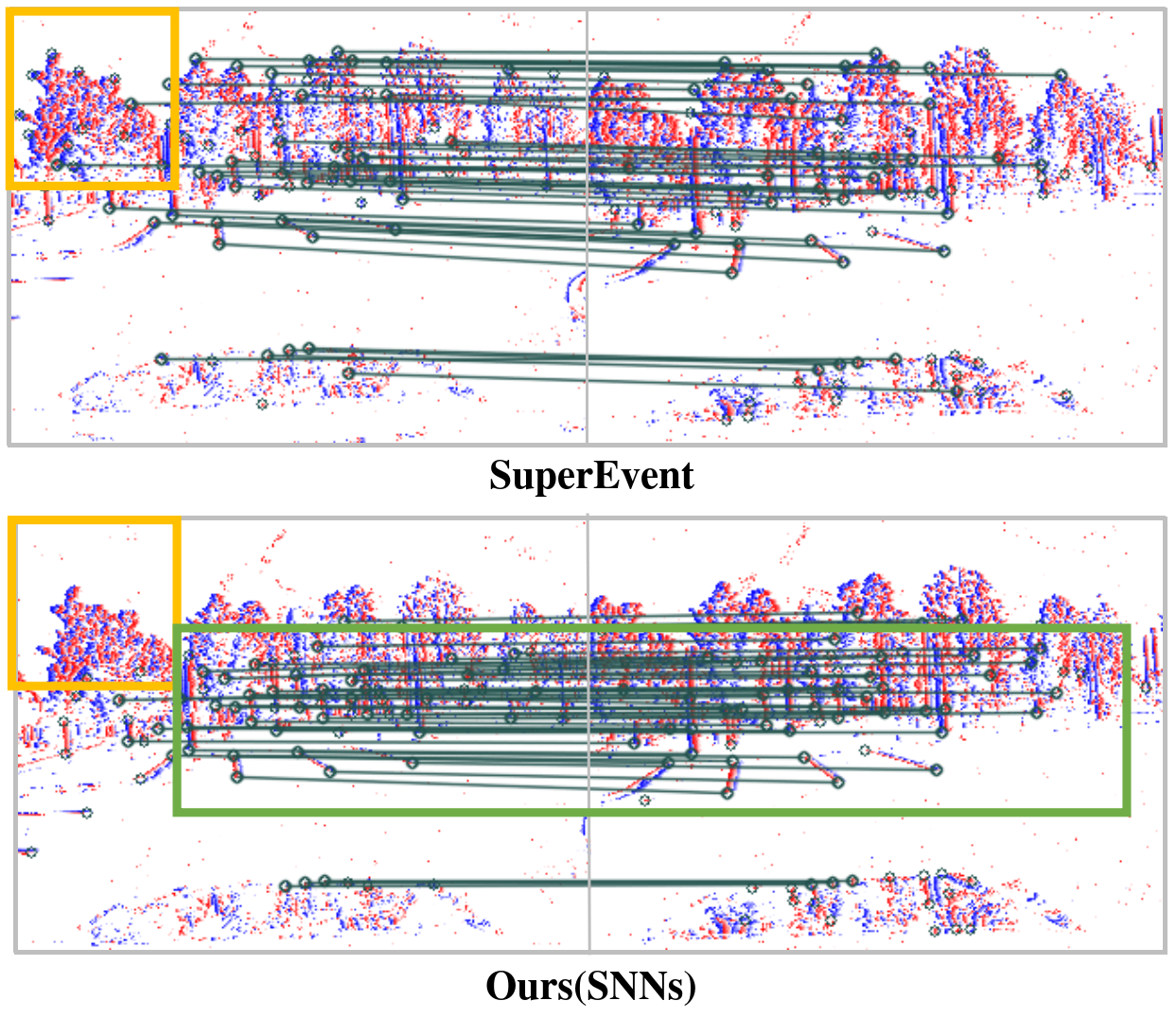}
    \caption{Qualitative comparison of feature detection and matching between the proposed method and the SuperEvent baseline. Yellow boxes highlight differences in keypoint detection, while green boxes show the improvements in matching density and spatial consistency brought by semantic fusion.}
    \label{fig:trajectory}
\end{figure}

Fig.~\ref{fig:trajectory} further compares the feature detection and matching results of our method with the SuperEvent~\cite{burkhardt2025superevent} baseline. The regions highlighted by the yellow boxes reveal clear differences in keypoint distribution between the two methods. The baseline produces numerous responses near the tree–sky boundary, whereas our method employs semantic modulation to enhance detection scores in geometrically stable regions. Consequently, reliable keypoints in these regions are preferentially retained during Top-K selection, while unstable responses near sky boundaries are suppressed. The tree–sky boundary in the upper part of the image further illustrates the difference in matching quality. The baseline produces a considerable number of spatially unstable matches in this region, whereas our method substantially reduces such erroneous correspondences. Meanwhile, within the semantically enhanced regions marked by the green boxes, our method yields denser matches with better spatial consistency, indicating that semantic information guides correspondences toward scene structures with well-defined geometric constraints. These results demonstrate that the performance improvement of our method does not arise from simply increasing the number of keypoints, but rather from the joint enhancement of keypoint localization reliability and descriptor discriminability. At the representation level, spike activations provide low-bit discrete encoding for event features; at the selection level, semantic modulation constrains the spatial distribution of detection responses. Through the coordinated optimization of feature representation and feature selection, the proposed framework enables the network to extract higher-quality keypoints and establish more reliable correspondences.

\subsection{VIO Trajectory Evaluation}
To further evaluate the practical value of the proposed method in downstream SLAM systems, we integrate the event-based feature methods into the stereo visual-inertial SLAM framework OKVIS2~\cite{leutenegger2022okvis2}. We then conduct VIO experiments on the TUM-VIE dataset~\cite{klenk2021tumvie}. Performance is measured by absolute trajectory error (ATE) in centimeters. SuperEvent~\cite{burkhardt2025superevent} reports integration results with OKVIS2. However, its interface module has not been released. To ensure a fair comparison, we implement the interface between the event features and OKVIS2. We rerun both SuperEvent and our method using the same pipeline. Therefore, all methods are evaluated under identical conditions.

\begin{figure*}[t]
    \centering
    \includegraphics[width=0.9\textwidth]{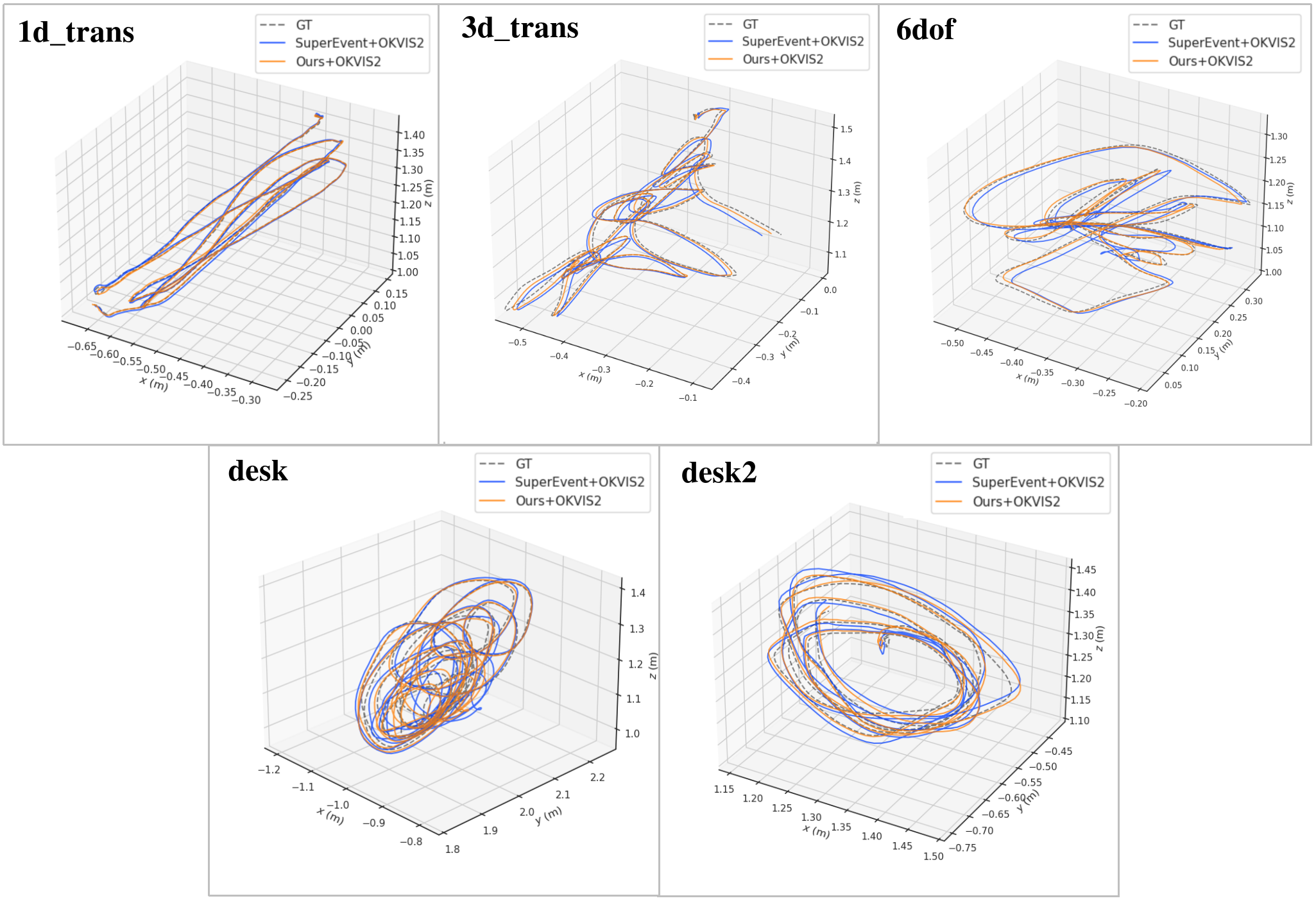}
    \caption{Trajectory comparison among our method (orange solid lines), the baseline method (blue solid lines), and the ground truth (gray dashed lines) on the TUM-VIE sequences~\cite{klenk2021tumvie}.}
    \label{fig:traj_compara}
\end{figure*}

Tab.~\ref{tab:vio} compares the ATE on the small-scale motion-capture sequences of the TUM-VIE dataset~\cite{klenk2021tumvie}. The results of existing event-based methods, including DEIO~\cite{guan2024deio} and ESVO~\cite{zhou2021esvo}, are taken from their original papers. SuperEvent~\cite{burkhardt2025superevent} and Ours (SNN) are evaluated using the same VIO pipeline to ensure a consistent comparison. Under this setting, Ours (SNN) achieves the lowest average ATE of 0.96~cm, corresponding to a 44\% reduction relative to the reproduced baseline, which obtains an average ATE of 1.71~cm. Fig.~\ref{fig:traj_compara} presents the trajectory estimation results on the TUM-VIE sequences~\cite{klenk2021tumvie} obtained by integrating OKVIS2~\cite{leutenegger2022okvis2} with our method and the baseline features, respectively. The orange solid lines denote the trajectories estimated using our method, the blue solid lines represent those obtained with the baseline, and the gray dashed lines indicate the ground-truth trajectories. As shown, the VIO system based on the proposed spiking features can stably track the camera pose, with the estimated trajectories remaining closely aligned with the ground truth. These results demonstrate that the proposed SNN-based spiking features can be reliably integrated into a complete VIO pipeline.
\begin{table*}[t]
    \centering
    \footnotesize
    \caption{ATE [cm] on TUM-VIE mocap sequences~\cite{klenk2021tumvie}.}
    \setlength\tabcolsep{3.5pt}
    \renewcommand{\arraystretch}{1.3}
    \label{tab:vio}
    \begin{tabular}{l c c c c c c c}
        \toprule
        Method & Modality & 1d-trans & 3d-trans & 6dof & desk & desk2 & Average \\
        \midrule
        DEIO~\cite{guan2024deio} & Mono E + IMU & 1.08 & 1.12 & 1.39 & 1.41 & 1.19 & 1.24 \\
        ESVO~\cite{zhou2021esvo} & Stereo E & 12.54 & 17.19 & 13.46 & 12.92 & 4.42 & 12.11 \\
        ES-PTAM~\cite{ghosh2024esptam} & Stereo E & 1.05 & 8.53 & 10.25 & 2.50 & 7.20 & 5.91 \\
        ICRA'24~\cite{niu2024imu} & Stereo E + IMU & 3.85 & 18.90 & failed & 8.99 & 9.47 & -- \\
        ESVO2~\cite{niu2024esvo2} & Stereo E + IMU & 3.33 & 7.26 & 3.21 & 6.16 & 4.02 & 4.78 \\
        OKVIS2~\cite{leutenegger2022okvis2} + SuperEvent~\cite{burkhardt2025superevent} (our impl.) & Stereo E + IMU & 0.49 & 1.88 & 2.25 & 2.07 & 1.87 & 1.71 \\
        \rowcolor{gray!20}
        \textbf{OKVIS2~\cite{leutenegger2022okvis2} + Ours (SNNs)} & Stereo E + IMU & \textbf{0.40} & \textbf{1.05} & \textbf{1.29} & \textbf{1.04} & \textbf{1.00} & \textbf{0.96} \\
        \bottomrule
    \end{tabular}
\end{table*}

\subsection{Inference Efficiency}
Tab.~\ref{tab:efficiency} summarizes the model size, inference speed, and theoretical energy consumption of each method. All evaluations are conducted at a resolution of $480\times640$ using FP32 precision.

\begin{table}[t]
    \centering
    \footnotesize
    \caption{Model Efficiency Comparison.}
    \setlength\tabcolsep{10pt}
    \renewcommand{\arraystretch}{1.4}
    \label{tab:efficiency}
    \begin{threeparttable}
        \begin{tabular}{l c c c}
            \toprule
            Methods & \#Params (M) & FPS & Power (mJ)\\
            \midrule
            SuperEvent (ANNs) & 1.54 & 110 & 7.50\\
            \rowcolor{gray!20}
            \textbf{Ours (ANNs)} & 2.39 & 93 & 12.02\\
            \rowcolor{gray!20}
            \textbf{Ours (SNNs)}  & 2.39 & 85 & 2.52\\
            \bottomrule
        \end{tabular}
        \begin{tablenotes}
            \footnotesize
            \item[*] All models are evaluated at $480\times640$ resolution with FP32 precision.
        \end{tablenotes}
    \end{threeparttable}
\end{table}

Compared with SuperEvent~\cite{burkhardt2025superevent}, Ours (ANN) introduces a semantic modulation module, increasing the number of parameters from 1.54 M to 2.39 M and reducing the inference speed from 110 FPS to 93 FPS. After further incorporating spiking activations, Ours (SNN) introduces no additional learnable parameters and therefore retains the same parameter count of 2.39 M, while achieving an inference speed of 85 FPS, slightly lower than that of its ANN counterpart. This reduction is mainly attributable to the lack of dedicated hardware support for spiking computation on general-purpose GPUs. Operations involved in spiking activations, such as rounding-based quantization and bit-plane decomposition, must still be simulated using floating-point computation, preventing the model from fully exploiting the sparse, event-driven, and asynchronous nature of SNNs. Therefore, the reported inference speed of Ours (SNN) primarily reflects its software simulation efficiency on a conventional GPU and should not be regarded as representative of its actual performance on a target neuromorphic platform.

The theoretical energy consumption is estimated using the operation-count-based method commonly adopted in SNN studies~\cite{kim2020spiking}. Owing to the additional computation introduced by the semantic modulation module, Ours (ANN) has a theoretical energy consumption of 12.02 mJ, slightly higher than the 7.50 mJ of the baseline. After spiking activations are introduced, part of the multiply–accumulate operations can be replaced with lower-cost accumulation operations, reducing the theoretical energy consumption of Ours (SNN) to 2.52 mJ. This corresponds to only 33.6\% of the baseline energy consumption and 20.9\% of that of the corresponding ANN model, yielding an approximately $4.8$-fold improvement in theoretical energy efficiency. It should be noted that this estimation is based on standard digital CMOS assumptions and does not account for additional overheads associated with membrane-potential updates and spike communication. Deployment on neuromorphic hardware, such as Loihi, may further improve energy efficiency by exploiting event-triggered asynchronous computation.

In summary, the proposed spiking inference model achieves accuracy comparable to that of its ANN counterpart while providing an approximately $4.8$-fold improvement in theoretical energy efficiency, demonstrating the energy-efficiency potential of spiking neural networks for event-based feature extraction and description.

\section{Conclusion}
This paper proposes E-S2Feat, the first framework to introduce spiking neural networks into event-based local feature detection and description. To accommodate the functional requirements and feature distributions of different network modules, we design spiking GELU and spiking SiLU activations and incorporate the existing MSF multi-level spiking activation. Together, these components form a module-specific spiking strategy that preserves local feature representation while improving inference efficiency. We further propose a semantic-guided feature modulation mechanism. Semantic information is used to adaptively modulate keypoint responses and enhance local descriptors, thereby encouraging the model to extract local features with greater geometric stability and stronger discriminability. Experimental results show that our model achieves competitive performance in pose estimation and visual-inertial odometry. It also maintains accuracy comparable to that of its ANN counterpart while achieving an approximately $4.8$-fold improvement in theoretical energy efficiency. These results demonstrate the potential of E-S2Feat as a viable solution for low-power event-based visual localization on resource-constrained platforms. Future work will focus on deploying and optimizing the proposed method on neuromorphic hardware.

\bibliographystyle{IEEEtran}
\bibliography{References}

\end{document}